\documentclass[twocolumn]{svjour3}         

\smartqed  

\usepackage[dvips]{graphicx}
\usepackage{natbib}
\usepackage{amsmath}
\usepackage{amsfonts}
\usepackage{mathptmx}
\usepackage{subfigure}

\title{Fatigue evaluation in maintenance and assembly operations by digital human simulation}
\author{Liang MA \and
        Damien CHABLAT \and
        Fouad BENNIS \and
        Wei ZHANG \and
        Bo HU \and
        Fran\c{c}ois GUILLAUME
}

\institute{Liang MA \and Damien CHABLAT \and Fouad BENNIS \at
               Institut de Recherche en Communications et Cybern\'{e}tique de Nantes, UMR 6597 du CNRS, \'Ecole Centrale de Nantes, IRCCyN - 1, rue de la No\"{e} - BP 92 101 - 44321 Nantes CEDEX 03, FRANCE \\
              Tel.:+33-02 40 37 69 58\\
              Fax: +33-02 40 37 69 30\\
              \email{\{liang.ma, damien.chablat, fouad.bennis\}@irccyn.ec-nantes.fr}
           \and
           Wei ZHANG \and Bo HU \at
              Department of Industrial Engineering, Tsinghua University, 100084, Beijing, P.R.CHINA\\
              \email{zhangwei@tsinghua.edu.cn, b-hu05@mails.tsinghua.edu.cn}
            \and
          Fran\c{c}ois GUILLAUME \at   EADS Innovation Works, 12, rue Pasteur - BP 76, 92152 Suresnes Cedex - FRANCE\\
          \email{francois.guillaume@eads.net}
}

\date{}

\begin{document}

\maketitle

\begin{abstract}
Virtual human techniques have been used a lot in industrial design in order to consider human factors and ergonomics as early as possible. The physical status (the physical capacity of virtual human) has been mostly treated as invariable in the current available human simulation tools, while indeed the physical capacity varies along time in an operation and the change of the physical capacity depends on the history of the work as well.
Virtual Human Status is proposed in this paper in order to assess the difficulty of manual handling operations, especially from the physical perspective. The decrease of the physical capacity before and after an operation is used as an index to indicate the work difficulty. The reduction of physical strength is simulated in a theoretical approach on the basis of a fatigue model in which fatigue resistances of different muscle groups were regressed from 24 existing maximum endurance time (MET) models. A framework based on digital human modeling technique is established to realize the comparison of physical status.
An assembly case in airplane assembly is simulated and analyzed under the framework. The endurance time and the decrease of the joint moment strengths are simulated. The experimental result in simulated operations under laboratory conditions confirms the feasibility of the theoretical approach.


\end{abstract}

\section{Introduction}
\label{intro}

Although automation techniques have played a very important role in industry, there are still lots of operations requiring manual handling operations thanks to the flexibility and the dexterity of human. Some of these manual handling operations deal with relative heavy physical loads, which might result in physical fatigue in the muscles and joints, and further generate potential risks for Musculoskeletal Disorders (MSDs) \citep{li1999cta}.

In order to improve the work design, digital human modeling (DHM) technique has been used more and more in industry taking human as the center of the work design system \citep{chaffin2002shr, chaffin2007hms}, since it benefits the validation of the workspace design, the assessment of the accessibility of an assembly design, the reduction of the production cost, and the reduction of the physical risks as well.

Several commercial available DHM tools have already been developed and integrated into computer aided design (CAD) tools, such as Jack \citep{BADLER1999}, 3DSSPP \citep{Chaffin1999}, RAMSIS \citep{bubb2006drp}, AnyBody \citep{damsgaard2006ams}, Santos$^{TM}$ \citep{VSR2004}, etc. In general, the virtual human in those tools is modeled with a large number of degrees of freedom (DOF) to represent the joint mobility, create the cinematic chain of human, and complete the skeleton structure of human. Meanwhile, the graphical appearance of virtual human is realized by bone, muscle, skin, and cloth models from the interior to the exterior, from simple stick models to complicated 3D mesh models. Normally, biomechanical database and anthropometry database are often set up to determine virtual human's dimensional and physical properties.

The main functions of the virtual human simulation tools are posture analysis and posture prediction. These tools are capable of determining the workspace of virtual human \citep{yang2008gaa}, assessing the visibility and accessibility of an operation \citep{Chedmail2003}, evaluating postures \citep{bubb2006drp}, etc. Conventional motion time methods (MTM) and posture analysis techniques can be integrated into virtual human simulation systems to assess the work efficiency \citep{Hou2007rvh}. From the physical aspect, the moment load of each joint (e.g., 3DSSPP) and even the force of each individual muscle (e.g., AnyBody) can be determined, and the posture is predictable for reach operations \citep{yang2006ppa} based on inverse kinematics and optimization methods. Overall, the human motion can be simulated and analyzed based on the workspace information, virtual human strength information, and other aspects. However, there are still several limitations in the existing virtual human simulation tools.

There is no integration of physical fatigue model in most of the human simulation tools. The physical capacity is often initialized as constant. For example, the joint strength is assigned as joint maximum moment strength in 3DSSPP, and the strength of each muscle is set proportional to its physiological cross section area (PSCA) in AnyBody. The physical capacity keeps constant in the simulation, and the fatigue effect along time is not considered enough. However, the change of the physical status can be experienced everyday by everyone, and different working procedures generate different fatigue effects. Furthermore, it has been reported that the motion strategy depends on the physical status, and different strategies were taken under fatigue and non-fatigue conditions \citep{chen2000cld, fuller2008pmc}. Therefore, it is necessary to create a virtual human model with a variable physical status for the simulation.

Some fatigue models have been incorporated into some virtual human tools to predict the variable physical strength. For example, Wexler's fatigue model \citep{Wexler20002} has been integrated into Santos$^{TM}$ \citep{Vignes2004}, and Giat's fatigue model \citep{Giat1993}  has been integrated based on Hill's muscle model \citep{hill1938hsa} in the computer simulation by \citet{KOMURA2000}. However, either the muscle fatigue model has  too many variables for ergonomic applications (e.g. Wexler's model), or there is no confidential physiological principle for the fatigue decay term \citep{xia2008tam} in the previous studies. It is necessary to find a simple fatigue model interpretable in muscle physiological mechanism for ergonomics applications.

In addition, some assessments in those tools provide indexes generated by traditional evaluation methods (e.g., Rapid upper limb assessment (RULA)). Due to the intermittent recording procedures of the conventional posture analysis methods, the evaluation result cannot analyze the fatigue effect in details. In this case, a new fatigue evaluation tool should be developed and integrated into virtual human simulation.

In order to assess the variable human status, a prototype of a digital human modeling and simulation tool  developed in OpenGL is presented in this paper. This human modeling tool is under a virtual environment framework involving variable physical status on the basis of a fatigue model. 

The structure of the paper is as follows. First, a virtual human model is introduced into the framework for posture analysis based on kinematic, dynamic, biomechanical, and graphical modeling. Second, the framework is presented with a new definition called Human Status. Third, the fatigue model and fatigue resistance for different muscle groups are introduced. At last, an application case European Aeronautic Defence \& Space (EADS) Company is assessed using this prototype tool under the framework with experimental validation.

\section{Digital human modeling}

\subsection{Kinematic modeling of virtual human}

In this study, the human body is modeled kinematically as a series of revolute joints. The Modified Denavit-Hartenberg (modified DH) notation system \citep{khalil02} is used to describe the movement flexibility of each joint. According to the joint function, one natural joint can be decomposed into 1 to 3 revolute joints. Each revolute joint has its rotational joint coordinate, labeled as $q_i$, with joint limits: the upper limit  $q_i^U$ and the lower limit $q_i^L$. A general coordinate $\mathbf{q}=[q_1,q_2,\ldots,q_n]$ is defined to represent the kinematic chain of the skeleton.

\begin{figure}[htp]
	\centering	\includegraphics[width=0.45\textwidth]{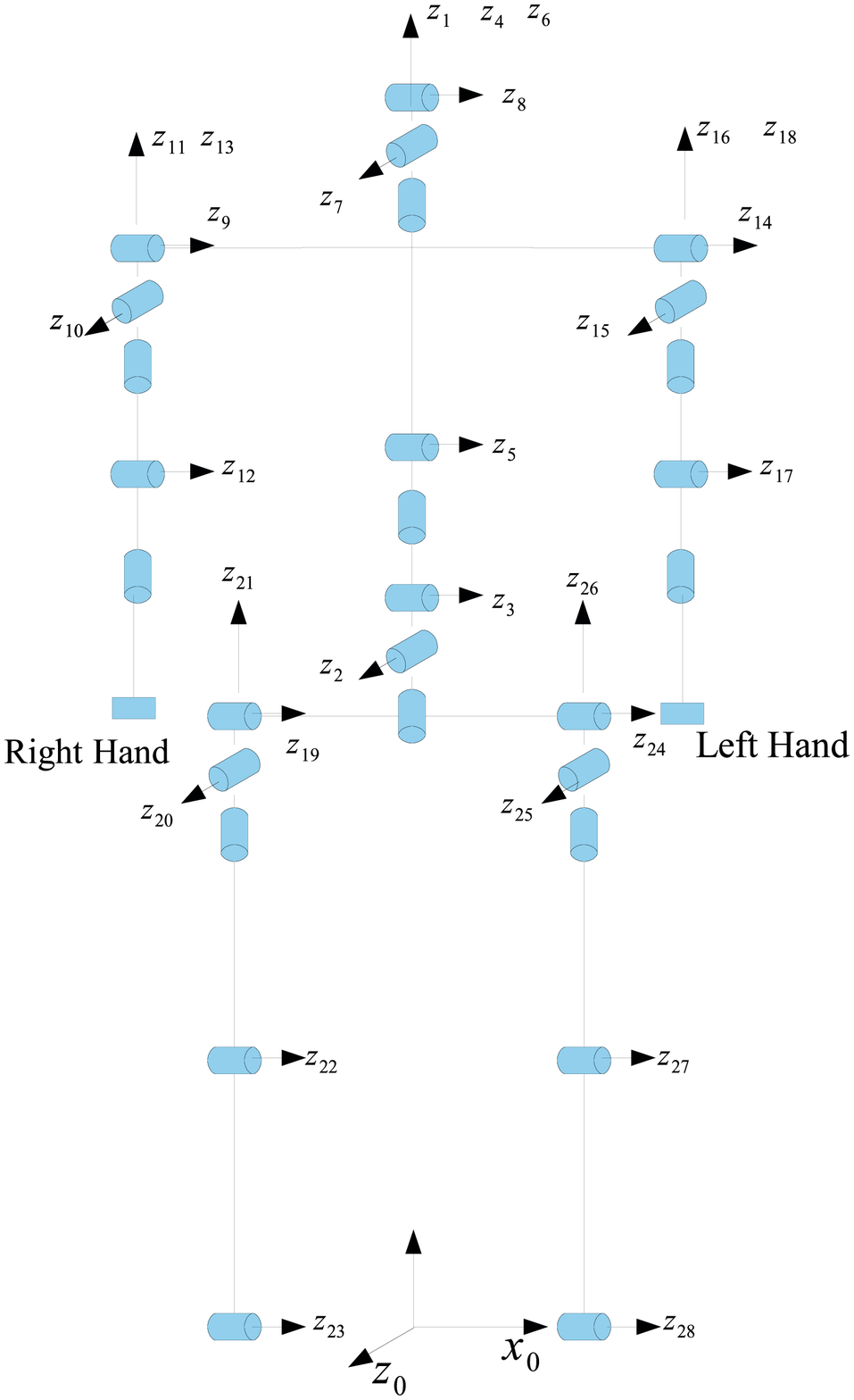}
	\caption{Geometrical modeling of virtual human}
	\label{fig:HumainStrucutre}
\end{figure}

The human body is geometrically modeled by 28 revolute joints to represent the main movement of the human body in Fig. \ref{fig:HumainStrucutre}. The posture, velocity, and acceleration are expressed by the general coordinates $\mathbf{q}$, $\dot{\mathbf{q}}$, and $\mathbf{\ddot{q}}$. It is feasible to carry out the kinematic analysis of the virtual human based on this kinematic model. By implementing inverse kinematic algorithms, it is able to predict the posture and trajectory of the human, particularly for the end effectors (e.g., the hands). All the parameters for modeling the virtual human are listed in Table \ref{tab:GeometricModelOverall}. $[X_r, Y_r, Z_r ]$ is the Cartesian coordinates of the root point (the geometrical center of the pelvis) in the coordinates defined by $X_0Y_0Z_0$.

\begin{table*}[ht]
	\centering
		\caption{Geometric modeling parameters of the overall human body}
		\begin{tabular}{ccccccccccc}
		\hline\noalign{\smallskip}
		$j$	& $a(j)$ & $u_j$ & $\sigma_j$ & $\gamma_j$ & $b_j$ & $\alpha_j$ & $d_j$ &  $q_j$ & $r_j$ &$q_{ini}$\\
		\hline\noalign{\smallskip}
		1	&	0	&	1	&0	&0	&$Z_r$&$-\frac{\pi}{2}$								&$X_r$		& $\theta_1$& $Y_r$&0\\
		
		2	&	1	&	0	&0	&0	&0		&$\frac{\pi}{2}$								&0				& $\theta_2$& 0&$\frac{\pi}{2}$\\
		3	&	2	&	0	&0	&0	&0		&$\frac{\pi}{2}$								&0				& $\theta_3$& 0&$\frac{\pi}{2}$\\
		4	&	3	&	0	&0	&0	&0		&$\frac{\pi}{2}$								&0				& $\theta_4$& $R_{lb}$&0\\
		5	&	4	&	0	&0	&0	&0		&$-\frac{\pi}{2}$								&0				& $\theta_5$& 0&0\\
		6 & 5 & 0	&0	&0	&0		&$\frac{\pi}{2}$								&0				& $\theta_6$& $R_{ub}$&$\frac{\pi}{2}$\\
		7 & 6 & 0	&0	&0	&0		&$\frac{\pi}{2}$								&0				& $\theta_7$& 0&$\frac{\pi}{2}$\\
		8 & 7 & 0	&0	&0	&0		&$\frac{\pi}{2}$								&0				& $\theta_8$& 0&0\\
		9 & 5 & 1	&0	&$-\frac{\pi}{2}$	&0		&0								&$D_{ub}$	& $\theta_9$& $-\frac{W_s}{2}$&0\\
		10& 9 & 0	&0	&0	&0		&$-\frac{\pi}{2}$								&0				& $\theta_{10}$& 0&$-\frac{\pi}{2}$\\
		11& 10 & 0	&0	&0	&0		&$-\frac{\pi}{2}$							&0				& $\theta_{11}$& $-R_{ua}$&$-\frac{\pi}{2}$\\
		
		12& 10& 0	&0	&0	&0		&$-\frac{\pi}{2}$								&0				& $\theta_{12}$& 0&0\\
		13& 11& 0	&0	&0	&0	&$\frac{\pi}{2}$									&0				& $\theta_{13}$& 0&0\\
		14& 5 & 1	&0	&$-\frac{\pi}{2}$	&0	&0									&$D_{ub}$	& $\theta_{14}$& $\frac{W_s}{2}$&0\\
		15& 14& 0	&0	&0	&0	&$-\frac{\pi}{2}$									&0				& $\theta_{15}$& 0&$-\frac{\pi}{2}$\\
		16& 15& 0	&0	&0	&0	&$-\frac{\pi}{2}$									&0				& $\theta_{16}$& $-R_{ua}$&$-\frac{\pi}{2}$\\
		17& 16& 0	&0	&0	&0	&$-\frac{\pi}{2}$									&0				& $\theta_{17}$& 0&0\\
		18& 17& 0	&0	&0	&0	&$\frac{\pi}{2}$									&0				& $\theta_{18}$& 0&0\\
		19& 1 & 1	&0	&$-\frac{\pi}{2}$	&0	&$-\frac{\pi}{2}$		&0				& $\theta_{19}$& $-\frac{W_w}{2}$&$-\frac{\pi}{2}$\\
		20& 19& 0	&0	&0	&0	&$-\frac{\pi}{2}$									&0				& $\theta_{20}$& 0&$-\frac{\pi}{2}$\\
		21& 20& 0	&0	&0	&0	&$-\frac{\pi}{2}$									&0				& $\theta_{21}$& $-R_{ul}$&$-\frac{\pi}{2}$\\
		22& 21& 0	&0	&0	&0	&$-\frac{\pi}{2}$									&0				& $\theta_{22}$& 0&$-\frac{\pi}{2}$\\
		23& 22& 0	&0	&0	&0	&0																&$-D_{ll}$& $\theta_{23}$& 0&0\\
		24& 1 & 1	&0	&$-\frac{\pi}{2}$	&0	&$-\frac{\pi}{2}$		&0				& $\theta_{24}$& $\frac{W_w}{2}$&$-\frac{\pi}{2}$\\
		25& 24& 0	&0	&0	&0	&$-\frac{\pi}{2}$									&0				& $\theta_{25}$& 0&$-\frac{\pi}{2}$\\
		26& 25& 0	&0	&0	&0	&$-\frac{\pi}{2}$									&0				& $\theta_{26}$& $-R_{ul}$&$-\frac{\pi}{2}$\\
		27& 26& 0	&0	&0	&0	&$-\frac{\pi}{2}$									&0				& $\theta_{27}$& 0&$-\frac{\pi}{2}$\\
		28& 27& 0	&0	&0	&0	&0																&$-D_{ll}$& $\theta_{28}$& 0&0\\
		\noalign{\smallskip}\hline
		\end{tabular}

	\label{tab:GeometricModelOverall}
\end{table*}

The geometrical parameters of the limb are required in order to accomplish the kinematic modeling. Such information can be obtained from anthropometry database in the literature. The dimensional information can also be used for the dynamic model of the virtual human. The lengths of different segments can be calculated as a  proportion of body stature $H$ in Table \ref{tab:BodySegmentLength}.

\begin{table}[htbp]
	\centering
	\caption{Body segment lengths as a proportion of body stature \citep{Chaffin1999,tilley2002mmw}}
		\begin{tabular}{llr}
		\hline\noalign{\smallskip}
		 Symbol&Segment&Length\\
		 \hline\noalign{\smallskip}
		 $R_{ua}$&Upper arm&0.186H\\
		 $R_{la}$&Forearm&0.146H\\
		 $R_{h}$&Hand&0.108H\\
		 $R_{ul}$&Thigh&0.245H\\
		 $D_{ll}$&Shank&0.246H\\
		 $W_{s}$&Shoulder width&0.204H\\
		 $W_{w}$&Waist width&0.100H\\
		 $D_{ub}, L_{ub}$&Torso length (L5-L1)&0.198H\\
		 $R_{ub}$&Torso length (L1-T1)&0.090H\\
			\hline\noalign{\smallskip}
		\end{tabular}
	\label{tab:BodySegmentLength}
\end{table}

\subsection{Dynamic modeling of virtual human}
Necessary dynamic parameters for each body segment include: gravity center, mass, moment of inertia about the gravity center, etc. According to the percentage distribution of total body weight for different segments \citep{Chaffin1999}, the weights of different segments can be calculated using Table \ref{tab:segmentweight}.

\begin{table}[htbp]
	\centering
	\caption{Percentage distribution of total body weight according to different segmentation plans  \citep{Chaffin1999}}
		\begin{tabular}{lr}
		\hline\noalign{\smallskip}
		Grouped segments,&individual segments\\
		\% of total body weight&\% of grouped-segments weight\\
		\hline\noalign{\smallskip}
		Head and neck=8.4\%&Head=73.8\% \\
		 & Neck=26.2\% \\
		 Torso=50\%&Thorax=43.8\%\\
		 &Lumbar=29.4\%\\
		 &Pelvis=26.8\%\\
		 Total arm=5.1\%&Upper Arm=54.9\%\\
		 &Forearm=33.3\%\\
		 &Hand=11.8\%\\
		 Total leg=15.7\%&Thigh=63.7\%\\
		 &Thigh=63.7\%\\
		 &Shank=27.4\%\\
		 &Foot=8.9\%\\
		\hline\noalign{\smallskip}
		\end{tabular}
	\label{tab:segmentweight}
\end{table}

It is feasible to calculate other necessary dynamic information with simplification of the segment shape. For limbs, the shape is simplified as a cylinder, head as a ball, and torso as a cube. The moment of inertia can be further determined based on the assumption of uniform density distribution. For the virtual human system, once all the dynamic parameters are known, it is possible to calculate the torques and forces at each joint following Newton-Euler method \citep{khalil02}.

\subsection{Biomechanical modeling of virtual human}
The biomechanical properties of the musculoskeletal system should also be modeled for virtual human simulation. From the physical aspect, the skeleton structure, muscle, and joint are the main biomechanical components in a human. In our study, only the joint moment strengths and joint movement ranges are used for the fatigue evaluation.

As mentioned before, with correct kinematic and dynamic models, it is possible to calculate torques and forces in joints with an acceptable precision. Although biomechanical properties of muscles are reachable and different optimization methods have been developed in the literature, the determination of the individual muscle force is still very complex and not as precise as that of joint torque \citep{xia2008tam}. Since there are several muscles attached around a joint, it creates an mathematical underdetermined problem for force calculation in muscle level. In addition, each individual muscle has different muscle fiber compositions, different levers of force, and furthermore different muscle coordination mechanisms, and the complexity of the problem will be increased dramatically in muscle level. Therefore, in our system, only the joint moment strength is taken to demonstrate the fatigue model.

\begin{figure}[h]
	\centering
		\includegraphics[width=0.45\textwidth]{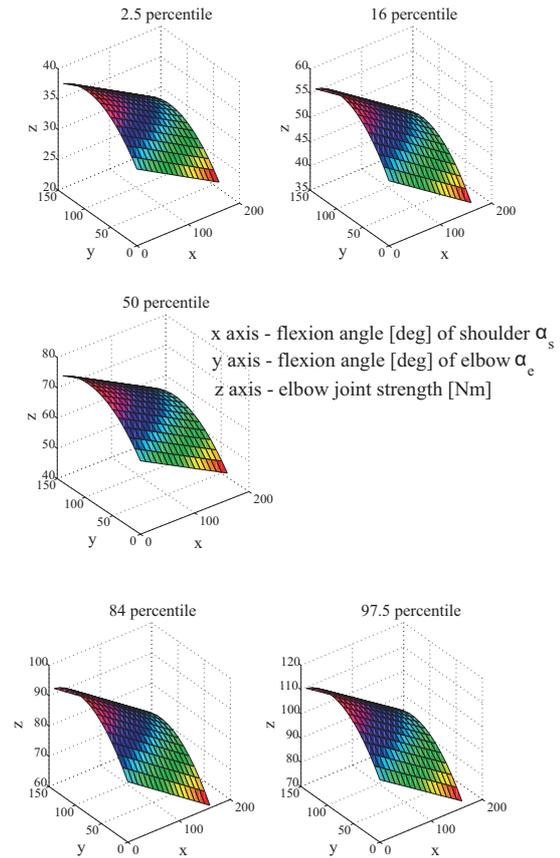}
	\caption{Elbow static strength depending on the human elbow and shoulder joint position, $\alpha_s$, $\alpha_e$ [deg]}
	\label{fig:Strength}
\end{figure}

The joint torque capacity is the overall performance of muscles attached around the joint, and it depends on the posture and the rotation speed of joint \citep{anderson2007mvj}. When a heavy load is handled in a manual operation, the action speed is relatively small, and it is almost equivalent to static cases. The influence from speed can be neglected, so only posture is considered. In this situation, the joint strength can be determined according to strength models in \cite{Chaffin1999}. The joint strength is measured in torque and modeled as a function of joint flexion angles. An example of joint strength is given in Fig. \ref{fig:Strength}. The shoulder flexion angle and the elbow flexion angle are used to determine the profile of the male adult elbow joint strength. The 3D mesh surfaces represent the elbow joint strengths for  95\% population. For the 50th percentile, the elbow joint strength varies from 45 to 75 N according to the joint positions.

\subsection{Graphical modeling of virtual human}
\begin{figure}[htp]
\centering
\subfigure[3DS model]
	{\label{fig:3ds}
	 \centering
	 \includegraphics[width=0.12\textwidth]{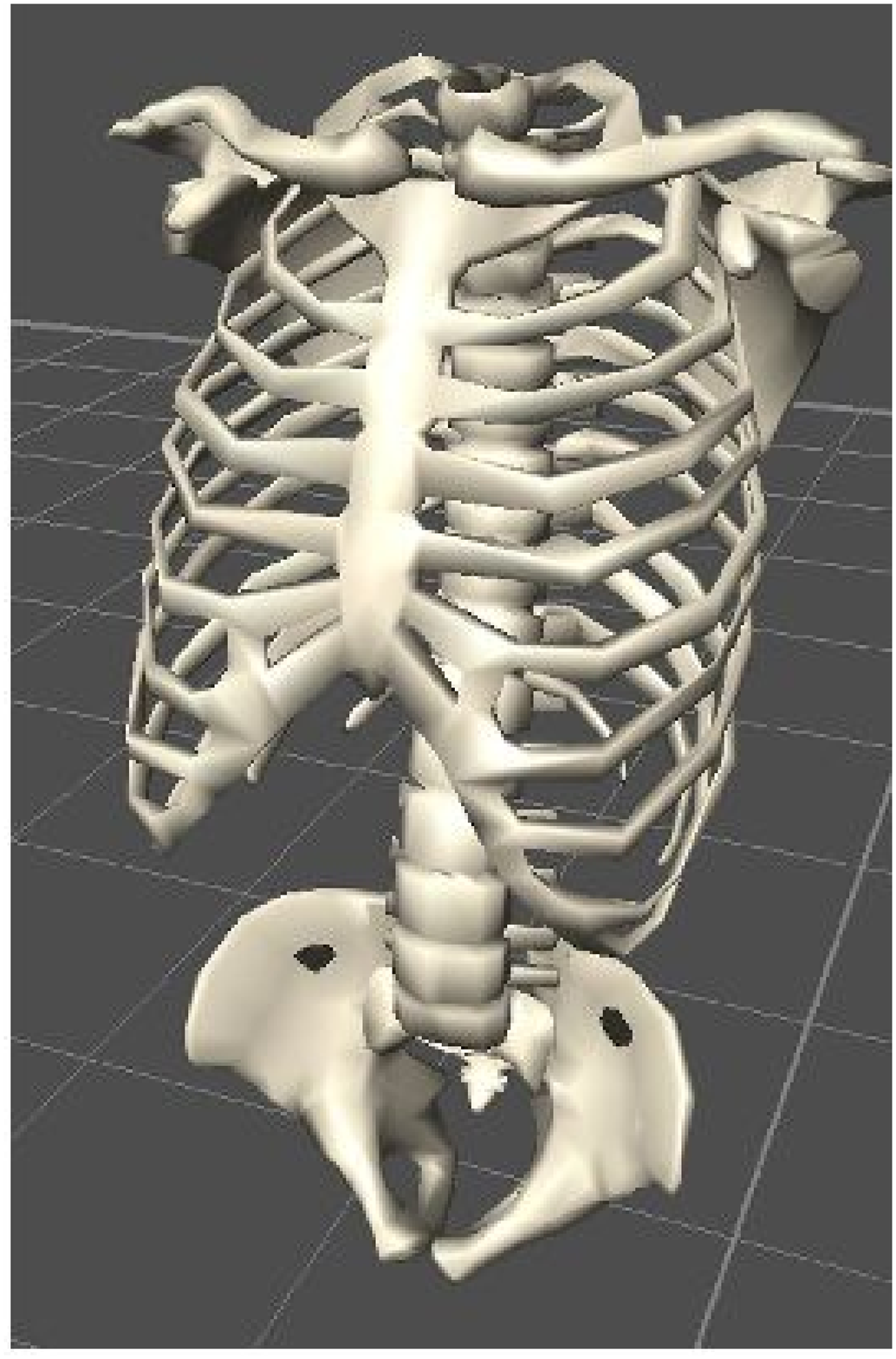}
	}\qquad
\subfigure[virtual skeleton]{\label{fig:skeleton}\includegraphics[width=0.18\textwidth]{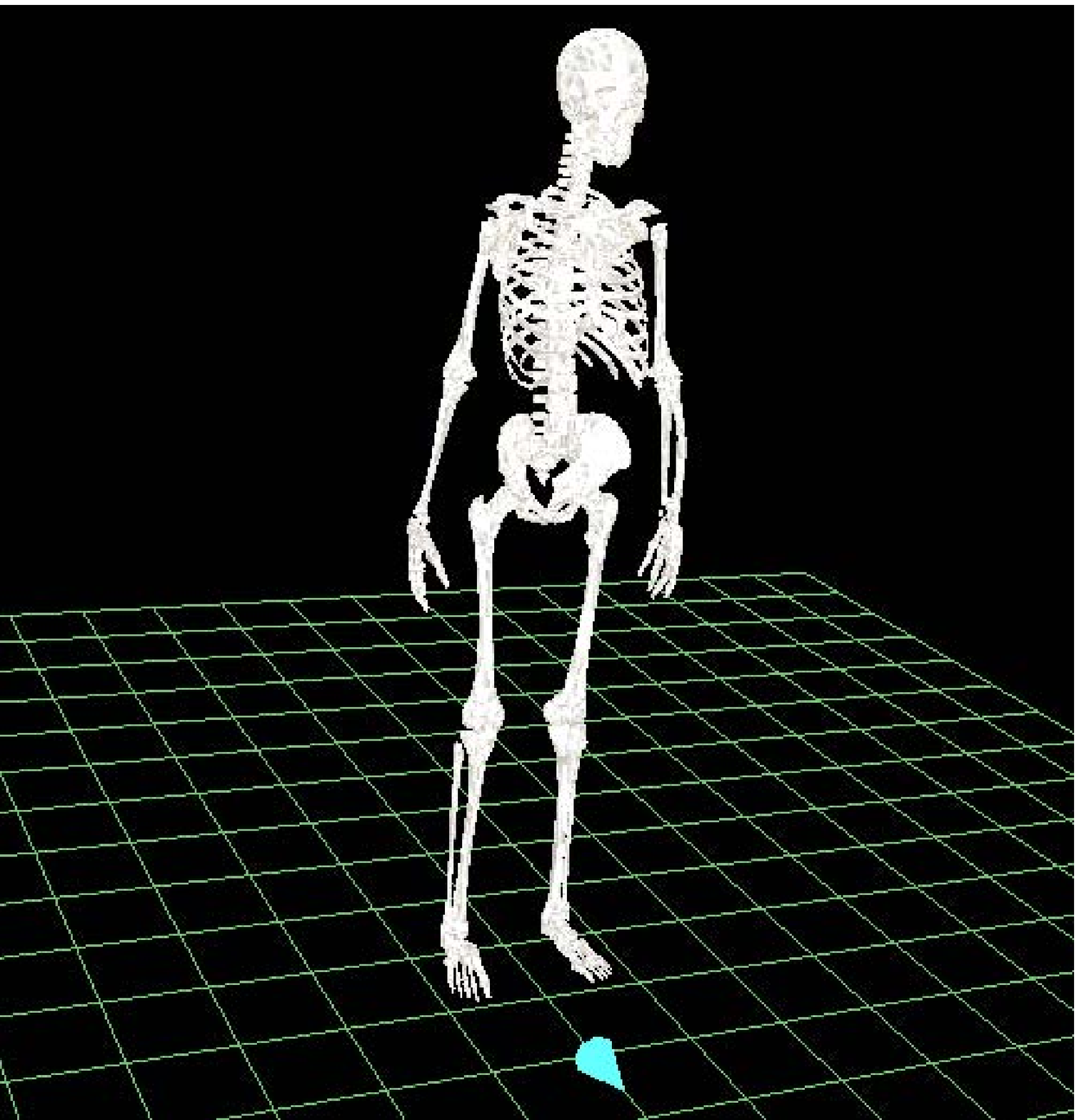}}\\
\caption{Virtual skeleton composed of 3DS models}
\end{figure}

The final step for modeling the virtual human is its graphical representation. The skeleton is divided into 11 segments:  body (1), head and neck (1), upper arms (2), lower arms (2), upper legs (2), lower legs (2), and feet (2). Each segment is modeled in 3ds file (3D Max, Autodesk Inc.) (Fig. \ref{fig:3ds}) and is connected via one or more revolute joints with another one to assemble the virtual skeleton (Fig. \ref{fig:skeleton}). For each segment, an original point and two vectors perpendicular to each other are attached to it to represent the position and the orientation in the simulation, respectively. The position and orientation can be calculated by the kinematic model of the virtual human.

\section{Framework for evaluating manual handling operations}
\label{sec:frame}

\begin{figure}[htp]
	\centering
		\includegraphics[width=0.45\textwidth]{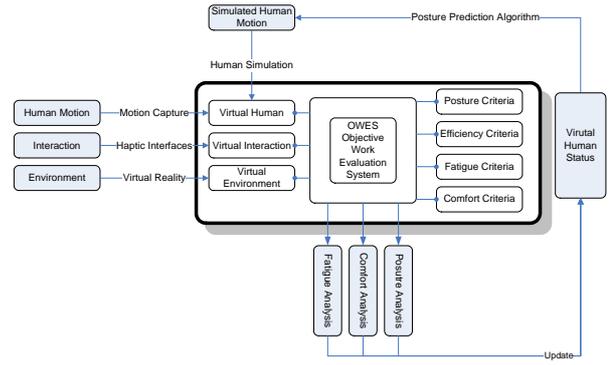}
	\caption{Framework for the work evaluation}
	\label{fig:ProjectEADS}
\end{figure}

The center of the framework is the objective work evaluation system (OWES) in Fig. \ref{fig:ProjectEADS}. The input module includes: human motion, interaction information, and virtual environment. Human motion is either captured by motion capture system or simulated by virtual human simulation. The interaction information is either obtained via haptic interfaces or modeled in simulation. Virtual environment is constructed to provide visual feedback to participants or workspace information in simulation. Input information is processed in OWES. With different evaluation criteria, different aspects of human work can be assessed as in the previous human simulation tools.

A new conception \textbf{human status} is proposed for this framework to generalize the discussion. \textbf{Human Status}: it is a state, or a situation in which the human possesses different capacities for an industrial operation. It can be further classified into mental status and physical status. Human status can be described as an aggregation of a set of human abilities, such as visibility, physical capacity (joint strength, muscle strength), and mental capacity. Virtual human status can be mathematically noted as $\mathbf{HS}=\{ \mathbf{V}_1,\mathbf{V}_2,\ldots, \mathbf{V}_n \}$. Each $\mathbf{V}_i$ represents one specific aspect of human abilities, and this state vector can be further detailed by a vector $\mathbf{V}_i=\{v_{i1},v_{i2},\ldots,v_{im_{i}}\}$. The change of the human status is defined as $\Delta\mathbf{HS}=\mathbf{HS}(t+\delta t)-\mathbf{HS}(t)=\{\Delta\mathbf{V}_1,\Delta\mathbf{V}_2,\ldots,\Delta\mathbf{V}_n \}$. For example, one aspect of the physical status can be noted as $\mathbf{HS}=\left[S_1, S_2, \ldots,S_n\right]$, where $S_i$ represents the physical joint strength of the $i^{th}$ joint of the virtual human.

In order to make the simulation as realistic as in real world, it is necessary to know how the human generates a movement. The bidirectional communication between human and the real world in an operation decides the action to accomplish a physical task: worker's mental and physical status can be influenced by the history of operation, while the worker chooses his or her suitable movement according to his or her current mental and physical statuses. Hence the framework is designed to evaluate the change of human status before and after an operation, and furthermore to predict the human motion according to the changed human status.

\begin{figure}[htp]
	\centering
		\includegraphics[width=0.45\textwidth]{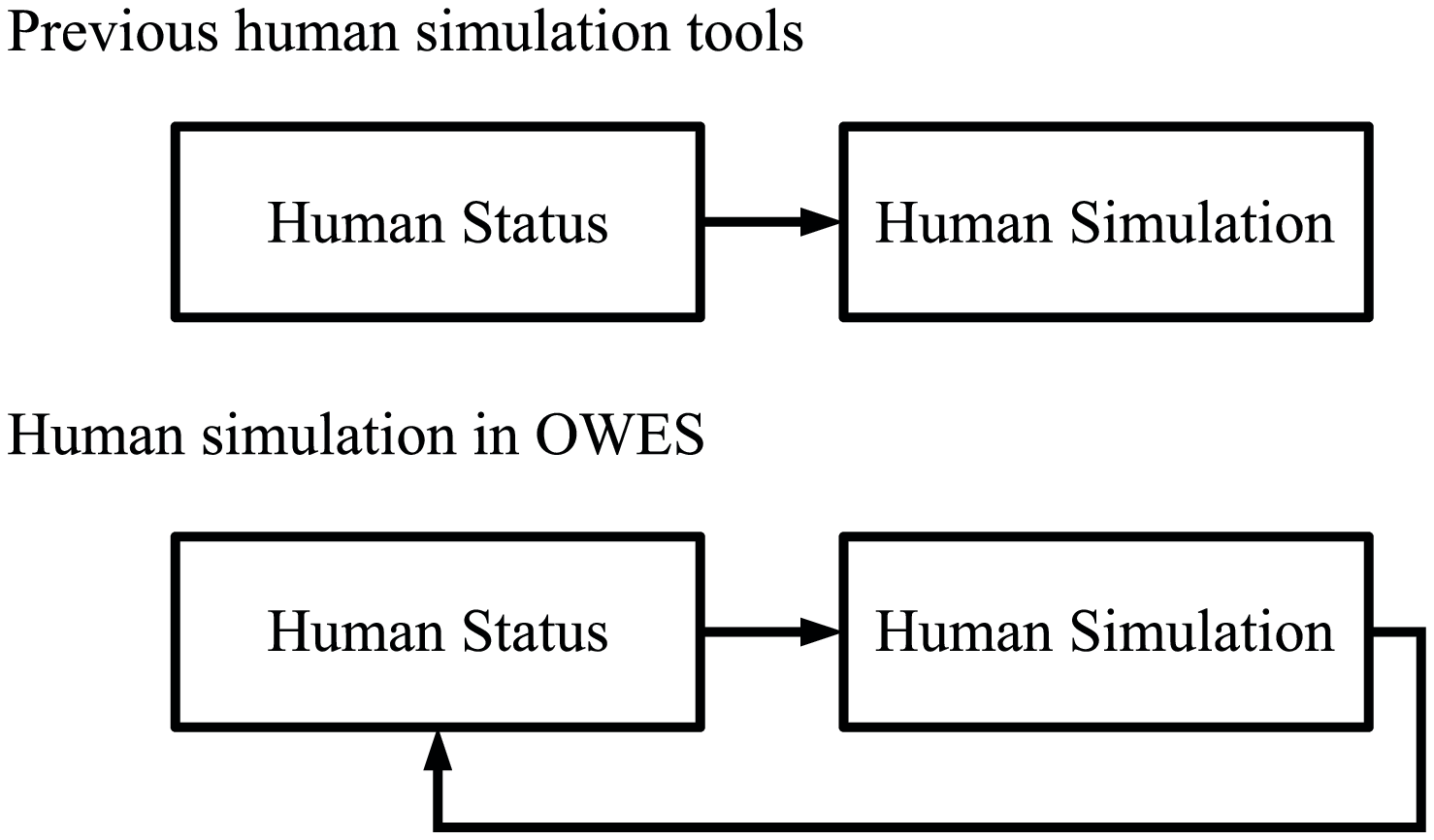}
	\caption{Human status in human simulation tools}
	\label{fig:HumanStatus}
\end{figure}

The human is often simplified  for posture control as a sensory-motor system in which there are enormous external sensors covering the human body and internal sensors in the human body capturing different signals, and the central nervous system (CNS) transfer the signals into decision making system (Cerebrum and Vertebral disc); the decision making system generates output commands to generate forces in muscles and then drives the motion and posture responding to the external stimulus. Normally, most of the external input information is directly measurable, such as temperature, external load, moisture, etc. However, how to achieve all the information for such a great number of sensors all over the human body is a challenging task. In addition, the internal perception of human body, which plays also an important role in motor sensor coordination, is much more difficult to be quantified. The most difficult issue is to know how the brain handles all the input and output signals while performing a manual operation. In previous simulation tools, the external input information has been already provided and handled. Visual feedback, audio feedback, and haptic feedback are often employed as input channel for a virtual human simulation. One limitation of the existing methods is that the internal sensation is not considered enough. Physical fatigue is going to be modeled and integrated into the framework to predict the perceived strength reduction and the reactions of the human body to the fatigue, which provides a close-loop for the human simulation (Fig. \ref{fig:HumanStatus}).

The special contribution in this framework is that the reduction of the physical strength can be evaluated in the framework based on a muscle fatigue model. And then the changed physical strength is taken as a feedback to the virtual human simulation to update the simulation result.

The framework performs mainly two functions: posture analysis and posture prediction (human simulation). The function of posture prediction is to simulate the human motion based on the current virtual human status.  Posture analysis focuses on assessing the difficulty of the manual operation. The difficulty of the work is assessed by the change of human status before and after the operation $\Delta\mathbf{HS}=\Delta \mathbf{HS}_{physical}$. Physical fatigue is one of the physical aspects, and this aspect is evaluated by the decrease of the strength in joints. The posture analysis function of the framework is our focus in this paper. 

More precisely in this paper,  the joint strengths models are used to determine the initial joint moment capacity, and then the fatigue in the joints can be further determined by the external load in the static operation and the fatigue model in Section \ref{sec:fatiguemodel}, and then the change of the physical status can be assessed.

\section{Fatigue modeling and Fatigue analysis}
\label{sec:fatiguemodel}
\subsection{Fatigue modeling}

A new dynamic fatigue model based on muscle motor unit recruitment principle was proposed in \citep{ma2008nsd}. This model was able to integrate task parameters (external load) and temporal parameters for predicting the fatigue of static manual handling operations in industry. Equation \ref{eq:FcemDiff} is the original form of the fatigue model to describe the reduction of the capacity. The descriptions of the parameters for Eq. \ref{eq:FcemDiff} are listed in Table \ref{tab:Parameters}. The detailed explanation about this model can be found in \citet{ma2008nsd}.

\begin{equation}
\label{eq:FcemDiff}
			\frac{dF_{cem}(t)}{dt} = -k \frac{F_{cem}(t)}{MVC}F_{load}(t)
\end{equation}

\begin{table}[htbp]
	\centering
	\caption{Parameters in dynamic fatigue model}
	\label{tab:Parameters}
		\begin{tabular}{lcp{0.3\textwidth}}
		\hline\noalign{\smallskip}
		Item & Unit & Description\\
		\hline\noalign{\smallskip}
		$MVC$					& $N$ &	Maximum voluntary contraction, maximum capacity of muscle\\
		$F_{cem}(t)$ 	& $N$ & Current exertable maximum force, current capacity of muscle\\
		$F_{load}(t)$	& $N$ & External load of muscle, the force which the muscle needs to generate\\
		$k$						& $min^{-1}$ & Constant value, fatigue ratio\\
		$\%MVC$				&				&Percentage of the voluntary maximum contraction\\
		$f_{MVC}$			&				&$\%MVC/100$, $\dfrac{F_{load}(t)}{MVC}$\\
		\hline\noalign{\smallskip}			
		\end{tabular}
\end{table}

Maximum endurance time (MET) models can be used to predict the endurance time of a static operation. In static cases, $F_{load}(t)$ is constant in the fatigue model, and then $MET$ is the duration in which $F_{cem}$ falls down to $F_{load}$. Thus, $MET$ can be determined in Eq. (\ref{eq:DynamicEndurance1}) and (\ref{eq:DynamicEndurance2}).

\begin{equation}
\label{eq:DynamicEndurance1}
	F_{cem}(t) = MVC \, e^{\int_{0}^{t} -k \dfrac{F_{load}(u)}{MVC}du} = F_{load}(t)
\end{equation}

\begin{equation}
\label{eq:DynamicEndurance2}
			t = MET = -\dfrac{ ln{\dfrac{F_{load}(t)}{MVC}}}{k\dfrac{F_{load}(t)}{MVC}} =-\dfrac{ ln(f_{MVC})}{k\, f_{MVC}}
\end{equation}

This model was validated in comparison with 24 MET models summarized in \citet{Khalid2006}. The previous MET models were used to predict the maximum endurance time for static exertions and they were all described in functions with $f_{MVC}$ as the only variable. High Pearson's correlations and interclass correlations (ICC) between the MET model in Eq. \ref{eq:DynamicEndurance2} and the other previous MET models validated the availability of our model for static cases. Meanwhile, the comparison between our model and a dynamic motor unit recruitment based model \citep{liu2002dmm} suggested that our model was also suitable for modeling muscle fatigue in dynamic cases.

In \citep{ma2008nsd}, the fatigue ratio $k$ was assigned 1 $min^{-1}$. However, from the literature, substantial variability in fatigue resistance in the population, and the variability results from several factors, such as age, career, gender, muscle groups, etc. The parameter $k$ can handle the effects on the fatigue resistance globally. Therefore, it is necessary to determine the fatigue resistances for different muscle groups to complete the muscle fatigue model.

\subsection{Fatigue resistance based on MET models}

Thanks to the high linear relationship between our MET model and the previous MET models, it is proposed that each static MET model $f(x)$ can be described mathematically by a linear equation (Eq. \ref{eq:linear}). In Eq. \ref{eq:linear}, $x$ is used to replace $f_{MVC}$ and $p(x)$ represents Eq. \ref{eq:DynamicEndurance2}. $\;m$ and $n$ are constants describing the linear relationship between static model and our model, and they need be determined in regression. Here, $m=1/k$ indicates the fatigue resistance of the static model, and $k$ is fatigue ratio or fatigability of different static model.

\begin{equation}
\label{eq:linear}
	f(x)=m\,p(x)+n
\end{equation}

Due to the asymptotic tendencies of MET models, when $x \to 1$ ($\%MVC \to 100$), $f(x) \to 0$ and $p(x) \to 0$ ($MET \to 0$), we assume $n=0$. Since some MET models are not suitable for $\%MVC\le15\%$, the regression is carried out from $x=0.16$ to $x=0.99$. With a step length 0.01, $N=84$ MET values are calculated to determine the parameter $m$ of each MET model by minimizing the function in Eq. \ref{eq:objective}.

\begin{equation}
\label{eq:objective}
	M(x)=\sum\limits_{i=1}^{N}\left(f(x_i)-m\,p(x_i)\right)^2=a\,m^2+b\,m+c
\end{equation}

From Eq. \ref{eq:objective}, $m$ can be calculated by Eq. \ref{eq:regression}.

\begin{equation}
\label{eq:regression}	m=\dfrac{-b}{2a}=\dfrac{\sum\limits_{i=1}^{N}p(x_i)f(x_i)}{\sum\limits_{i=1}^{N}p(x_i)^2}\,>0
\end{equation}

\begin{figure}[htp]
	\centering
		\includegraphics[width=0.45\textwidth]{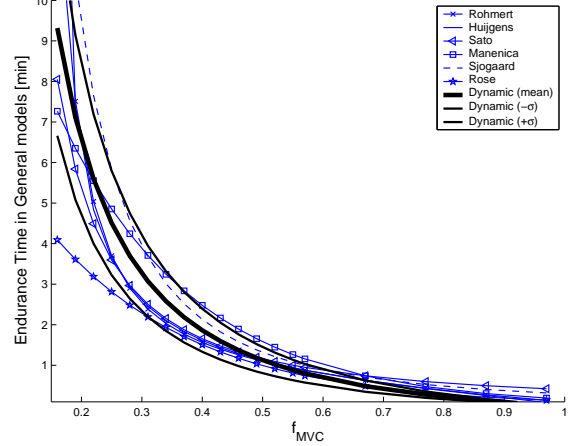}
	\caption{Prediction of MET in the dynamic MET model in comparison with that in the general models}
	\label{fig:generalresult}
\end{figure}

The regression result represents the fatigue resistance of the muscle group. In comparison with 6 general MET models, 6 elbow models, 5 shoulder models, and 6 hip/back models, different muscle fatigue resistances for corresponding muscle groups were calculated and listed below in Table \ref{tab:FatigueResistanceTable}. The mean value $\bar{m}$ and $\sigma_m$ can be  used to adjust our MET model to cover different MET models, and they can be further used to predict the fatigue resistance of a muscle group for a given population.  The prediction with mean value and its deviation in general MET models is shown in Fig.\ref{fig:generalresult}. It is observable that the bold solid curve and two slim solid curves cover most of the area formed by the previous empirical MET models.

It should be noted that the fatigue resistance for different muscle groups are only regressed based on the empirical data grouped in the literature, and the results (Table \ref{tab:ShoulderMETModels}) for shoulder and hip/back muscle groups did not conform to the normal distribution. For the shoulder joint,  the subjects in these models  were not only from different careers but also from different gender mixture. Therefore, the fatigue resistance result can only provide a reference in this study.

\begin{table}[htbp]
	\centering
	\caption{Fatigue resistance $\bar{m}$ for different muscle groups}
		\begin{tabular}{lccc}
		\hline\noalign{\smallskip}
		Segment&&$\bar{m}$&$\sigma_m$\\
		\hline\noalign{\smallskip}
		General&&0.8135\;& 0.2320\\
		Shoulder&&0.7562\;&0.4347\\
		Elbow&&0.8609\;&0.4079\\
		Hip&&1.9701\;&1.1476\\
		\hline\noalign{\smallskip}			
		\end{tabular}
	\label{tab:FatigueResistanceTable}
\end{table}

\begin{table}[htbp]
	\centering
	\caption{Fatigue resistances of shoulder MET models }
	\label{tab:ShoulderMETModels}
		\begin{tabular}{p{0.18\textwidth}p{0.2\textwidth}c}
		\hline Model&Subjects&$m$\\
		\hline
		\citet{sato1984eta}&5 male &0.427\\
		\citet{rohmert1986ssn}&6 male and 1 female students&0.545\\
		\citet{mathiassen1999psf}&20 male and 20 female municipal employees&0.698\\
		\citet{Garg2002}&12 female college subjects&1.393\\
		\hline	
		\end{tabular}
\end{table}

\subsection{Workflow for fatigue analysis}

\begin{figure}[htbp]
	\centering
		\includegraphics[width=0.45\textwidth]{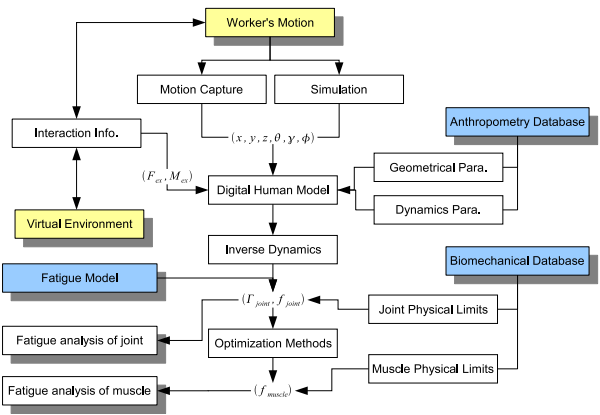}
	\caption{Workflow for the fatigue evaluation}
	\label{fig:workflow}
\end{figure}

The general process of the posture analysis has been discussed in Section \ref{sec:frame}, and here is the flowchart in Fig. \ref{fig:workflow} to depict all the details in processing all the input information.

First, human motion obtained either from human simulation or from motion capture system is further processed to displacement $\mathbf{q}$, speed $\dot{\mathbf{q}}$, and acceleration $\mathbf{\ddot{q}}$ in general coordinates.

The external forces and torques on the human body are either measured directly by force measurement instruments or estimated in the simulation. The external loads are transformed to $\Gamma_i$ and $F_i$ in the coordinates attached to $q_i$ in the modified DH method.

Human motion and interaction (forces, torques) are mapped into the digital human model which is geometrically and dynamically modeled from anthropometry database and the biomechanical database. Inverse dynamics is used to calculate the torque and force at each general joint. If it goes further, the effort of each individual muscle can be determined using optimization method as well.

Once the loads of the joints are determined, the fatigue of each joint can be analyzed using the fatigue model. The reduction of the physical strength can be evaluated, and finally the difficulty of the operation can be estimated by the change of physical strengths.

\section{Analysis Results for EADS Application Cases - Drilling}
\subsection{Operation description}
\begin{figure}[htp]
	\centering
		\includegraphics[width=0.45\textwidth]{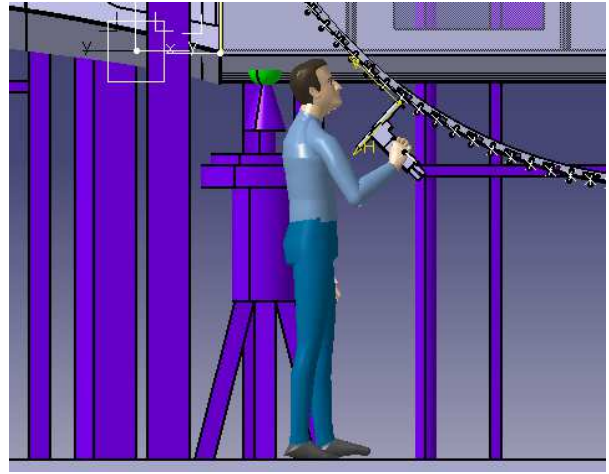}
	\caption{Drilling case in CATIA}
	\label{fig:drilling}
\end{figure}

The application case is the assembly of two fuselage sections with rivets from the assembly line of an airplane in European Aeronautic Defence \& Space (EADS) Company. One part of the job consists of drilling holes all around the cross section. The tasks is to drill holes around the fuselage circumference. The number of the holes could be up to 2,000 on an orbital fuselage junction of an airplane. The drilling machine has a weight around 5 kg, and even up to 7 kg in the worst condition with consideration of the pipe weight. The drilling force applied to the drilling machine is around 49N. In general, it takes 30 seconds to finish a hole. The drilling operation is illustrated in Fig. \ref{fig:drilling}. The fatigue happens often in shoulder, elbow, and lower back because of the heavy load. Only the upper limb is taken into consideration in this demonstration case to decrease the complexity of the analysis.

\subsection{Endurance time prediction}
The drilling machine with a weight 5 kg is taken to calculate the maximum endurance time under a static posture with shoulder flexion as $30^{\circ}$ and elbow flexion $90^{\circ}$ for maintaining the operation in a continuous way. The weight of the drilling machine is divided by two in order to simplify the load sharing problem. The endurance result is shown in Table \ref{tab:MET} for the population falling in the 95\% strength distribution. It is found that the limitation of the work is determined by the shoulder, since the endurance time for the shoulder joint is much shorter than that of the elbow joint.

\begin{table}[htbp]
	\caption{Maximum endurance time of shoulder and elbow joints for drilling work}
	\centering
		\begin{tabular}{cccccc}
			\hline\noalign{\smallskip}
			MET [sec]&$S-2\sigma$&$S-\sigma$&$S$&$S+\sigma$&$S+2\sigma$\\
			\hline\noalign{\smallskip}
			Shoulder\\
			\hline\noalign{\smallskip}
			$\bar{m}-\sigma_m$&19.34&45.05&75.226&108.81&145.16\\
			$\bar{m}$&45.489&105.96&176.94&255.94&341.44\\
			$\bar{m}+\sigma_m$&71.639&166.87&278.65&403.07&537.71\\
			\hline\noalign{\smallskip}
			Elbow\\
			\hline\noalign{\smallskip}
			$\bar{m}-\sigma_m$&230.61&424.27&640.47&873.52&1120.1\\
			$\bar{m}$&438.27&806.3&1217.2&1660.1&2128.6\\
			$\bar{m}+\sigma_m$&645.92&1188.3&1793.9&2446.6&3137.2\\
			\hline\noalign{\smallskip}
		\end{tabular}
	\label{tab:MET}
\end{table}

\begin{figure}[htp]
\centering
\includegraphics[width=0.45\textwidth]{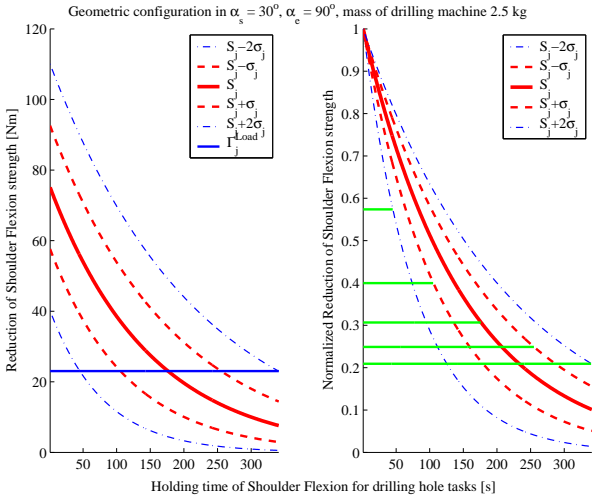}
\caption{Endurance time prediction for shoulder with average fatigue resistance}
\label{fig:shoulderadjusted}
\end{figure}

\begin{figure}
\centering
\includegraphics[width=0.45\textwidth]{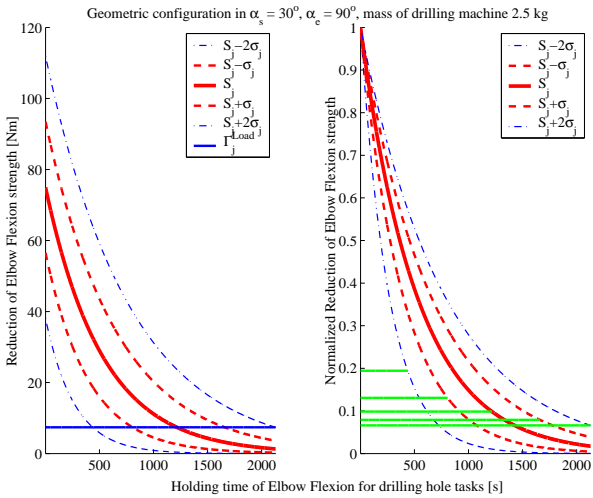}
\caption{Endurance time prediction for the elbow with average fatigue resistance}
\label{fig:elbowadjusted}
\end{figure}

\begin{figure}[htp]
\centering
\includegraphics[width=0.45\textwidth]{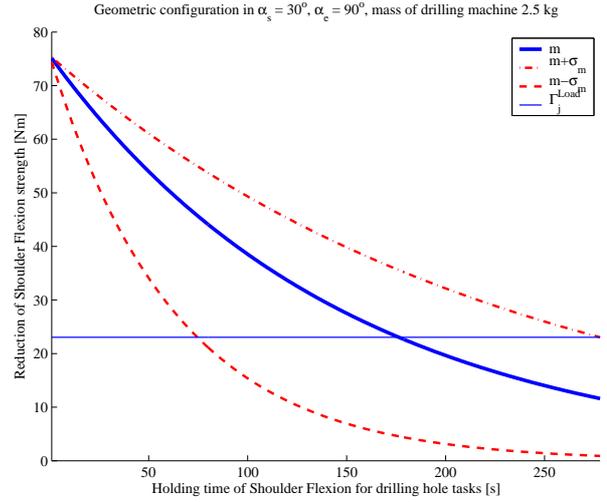}
\caption{Endurance time for the population with average strength  for  shoulder joint}
\label{fig:shouldererror}
\end{figure}

\begin{figure}
\centering
\includegraphics[width=0.45\textwidth]{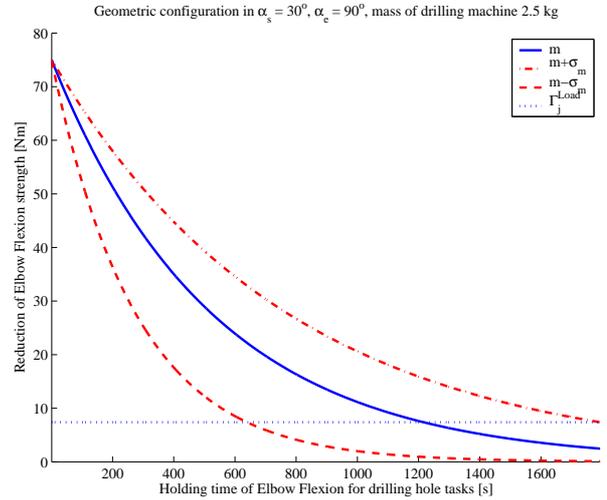}
\caption{Endurance time for the population with average strength  for  elbow joint}
\label{fig:METwithError}
\end{figure}

The difference in endurance results has two origins. One is the external load relative to the joint strength. The second comes from the fatigue resistance difference among the population. These differences are graphically presented from Fig. \ref{fig:shoulderadjusted} to Fig. \ref{fig:METwithError}. Figure \ref{fig:shoulderadjusted} and Figure \ref{fig:elbowadjusted} show the variable endurance caused by the joint strength distribution in the adult male population with the mean fatigue resistance. Larger strength results in longer endurance time for the same external load. Figure \ref{fig:shouldererror} and Figure \ref{fig:METwithError} present the endurance time for the population with the average joint strength but different fatigue resistances, and it shows that larger fatigue resistance leads to longer endurance time.  Combining with the strength distribution and the fatigue resistance variance, the MET can be estimated for all the population.

\subsection{Fatigue evaluation}
\begin{figure}[htp]
\centering	\includegraphics[width=0.45\textwidth]{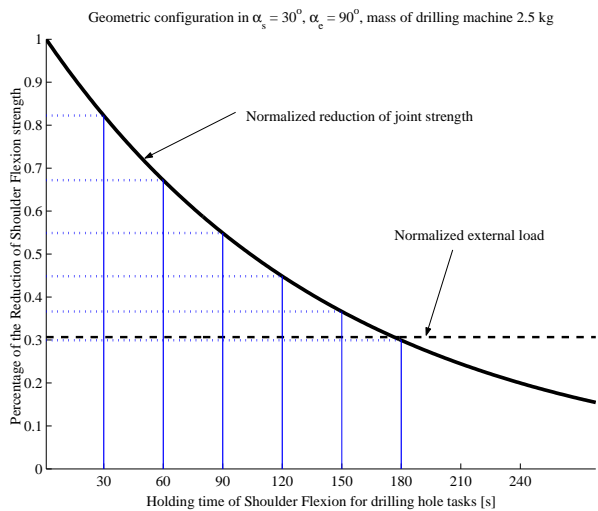}
	\caption{Fatigue evaluation after drilling a hole in a continuous drilling process}
	\label{fig:changeofpercentage}
\end{figure}

The fatigue is evaluated by the change of the joint strength in a fatigue operation. The working history can generate influence on the fatigue. Therefore, the fatigue for drilling a hole is evaluated in a continuous working process up to 6 holes. Only the population with the average strength and the average fatigue resistance is analyzed in fatigue evaluation in order to present the effect of the work history. The reduced strength is normalized by dividing the maximum joint strength, and it is shown in Fig. \ref{fig:changeofpercentage}. It takes 30 seconds to drill a hole, and the joint strength is calculated and normalized every 30 seconds until exhaustion for the shoulder joint.

In our current research, $\mathbf{HS}$ includes only the joint strength vector. The evaluation of the fatigue is measured by the change of the joint strength for drilling a hole. The result is shown in Table \ref{tab:Normalized}. Three measurements are given in this table: one is the normalized physical strength every 30 seconds, noted as $\dfrac{\mathbf{HS}_i}{\mathbf{HS}_{max}}$; one is the difference between the joint strength before and after finishing a hole, noted as $\dfrac{\mathbf{HS}_{i}-\mathbf{HS}_{i+1}}{\mathbf{HS}_{max}}$; the last one if the difference between the joint strength and the maximum joint strength, noted as $\dfrac{\mathbf{HS}_{max}-\mathbf{HS}_{i}}{\mathbf{HS}_{max}}$.
In Table \ref{tab:Normalized}, only the reduction of the shoulder joint strength is presented, since the relative load in elbow joint is much smaller.

\begin{table*}[htbp]
	\centering
	\caption{Normalized shoulder joint strength in the drilling operation}		
	\begin{tabular}{lccccccc}
	\hline Time [s]&0&30&60&90&120&150&180\\
	\hline\noalign{\smallskip}	$\bar{m}$\\
\hline\noalign{\smallskip}
$\dfrac{\mathbf{HS}_i}{\mathbf{HS}_{max}}$&100\%&82.2\%&67.2\%&54.9\%&44.8\%&36.6\%&30.1\%\\ \noalign{\smallskip}$\dfrac{\mathbf{HS}_{i}-\mathbf{HS}_{i+1}}{\mathbf{HS}_{max}}$&0\%&17.8\%&15.0\%&12.3\%&10.1\%&8.2\%&6.5\%\\
\noalign{\smallskip}$\dfrac{\mathbf{HS}_{max}-\mathbf{HS}_{i}}{\mathbf{HS}_{max}}$&0\%&17.8\%&32.8\%&45.1\%&55.2\%&63.4\%&69.9\%\\
\hline\noalign{\smallskip}
	\end{tabular}
	\label{tab:Normalized}
\end{table*}

From Fig. \ref{fig:changeofpercentage} and Table \ref{tab:Normalized}, the joint strength keeps the trend of descending in the continuous work. The ratio of the reduction gets smaller in the work progress due to the physiological change in the muscle fiber composition. More time consumed to work leads more reduction in physical strengths. The reduction relative to the maximum strength is able to assess the difficulty of the operations.

\subsection{Experiment validation}
Simulated drilling operations were tested under laboratory conditions in Tsinghua University. A total of 40 male industrial workers were asked to simulate the drilling work in a continuous operation for 180 seconds. Maximum output strengths were measured in the simulated operations at different periods of the operation. Fatigue was indexed by the reduction of the joint strength along time relative to the initial maximum joint strength. Three out of the 40 subjects could not sustain the external load for a duration of 180 seconds, and 34 subjects had a shoulder joint fatigue resistance (Mean=1.32, SD=0.62) greater than the average shoulder joint fatigue resistance in Table \ref{tab:FatigueResistanceTable}, which means that the sample population has a higher fatigue resistance than the population grouped in the regression.

The physical strength has been measured in simulated job static strengths, and the reduction in the operation varies from 32.0\% to 71.1\% (Mean=53.7\% and SD=9.1\%). The reduction falls in the fatigue prediction of the theoretical methods in Table \ref{tab:fatigue} (Mean=51.7\%, SD=12.1\%).

\begin{table}[h]
	\caption{Normalized torque strength reduction for the population with higher fatigue resistance}
	\centering
		\begin{tabular}{lccccc}
			\hline\noalign{\smallskip}			 $\dfrac{\mathbf{HS}_{max}-\mathbf{HS}_{180}}{\mathbf{HS}_{max}}$&$S-2\sigma$&$S-\sigma$&$S$&$S+\sigma$&$S+2\sigma$\\
			\hline\noalign{\smallskip}	
			$\bar{m}$&-&-&69.9\%&62.5\%&56.3\%\\
			$\bar{m}+\sigma_m$&-&63.2\%&53.2\%&46.4\%&40.8\%\\
			$\bar{m}+2\sigma_m$&64.9\%&51.9\%&43.0\%&36.7\%&31.9\%\\
			\hline\noalign{\smallskip}
		\end{tabular}
	\label{tab:fatigue}
\end{table}

\subsection{Discussion}
Under the proposed framework, the conception of the virtual human status is introduced and realized by a virtual human modeling and simulation tool. The virtual human is kinematic modeled based on the modeling method in robotics. Inverse dynamics is used to determine the joint loads. With the integration of a general fatigue model, the physical fatigue in a manual handling operation in EADS is simulated and analyzed. The decrease in human joint strengths can be predicted in the theoretical approach, and it has been validated with experimental data.

Human status is introduced in this framework in order to generalize all the discussion for the human simulation. We concentrate only on the physical aspect of the virtual human, in particular on joint strengths. Physical status can be extended to other aspects, either measurable using instruments (e.g., heart rate, oxygen consumption, electromyograph of muscle, etc.) or predicable using mathematical models (e.g., vision, strength, etc.). Similarly, the mental status of human can also be established by similar terms (e.g., mental capacity, mental workload, mental fatigue, etc.). Under the conception of human status, different aspects of the human can be aggregated together to present the virtual human completely. The changed human status caused by a physical job or a mental job can be measured or predicted to assess different aspects of the job. It should be noted that the definition of human status is still immature and it requires great effort to form, extend, and validate this conception.

The main difference between the fatigue analysis in our study and the previous methods for posture analysis is: in previous methods \citep{wood1997mfd,iridiastadi2006emfa,romanliu2005dfc}, intermittent procedures were used to develop the fatigue model with job specific parameters; in contrast, all the related physical exposure factors are taken into consideration in a continuous approach in our model. In this way, the analysis of the manual handling operation can be generalized without limitations of job specific parameters. Furthermore, the fatigue and recovery procedures can be decoupled to simplify the analysis in a continuous way. Although only a specific application case is presented in this paper, the feasibility of the general concept has been verified by the introduction of human status and the validation of the fatigue model.

It should be noted that the recovery of the physical strength has not been considered yet. Although there are several work-rest allowance models in the literature, substantial variability was found among the prediction results for industrial operations \citep{elahrache2009cra} and it is still ongoing to develop a general recovery model.

\section{Conclusions}
In this study, human status is introduced into the work evaluation system, especially for the physical status. It provides a global definition under which different aspects of human abilities can be integrated and assessed simultaneously. The effect of the work on the human status, either positive or negative, can be measured by the change of the human status before and after the operation. We concentrate our study on physical aspects, especially on joint moment strengths.
The physical fatigue analysis in a drilling case under the work evaluation framework demonstrates the work flow and the functions of the virtual human simulation. The change of joint moment strength, a specific aspect of human physical status, has been simulated based on a general fatigue model with fatigue resistances. The similar results between the analysis and the experimental data suggests that the framework may be useful for assessing the physical status in continuous static  operations.

The new conception human status and the theoretical method for assessing the physical status may provide a new approach to generalize the virtual human simulation and evaluate the physical aspect in continuous static manual handling operations. This approach is useful to assess the physical load to prevent industrial workers from MSD risks, and it can also be used to assess mental load with extension of mental status.

However, it should be noted that great effort has to be done to extend different aspects in human status to make it more precise. Even only for physical fatigue, it is still necessary to develop a recovery model to complete the fatigue prediction.

\section*{Acknowledgments}

This research was supported by the EADS and the R\'{e}gion des Pays de la Loire (France) in the context of collaboration between the \'{E}cole Centrale de Nantes (Nantes, France) and Tsinghua University (Beijing, PR China).
\bibliographystyle{spbasic}

\end{document}